\documentclass[sigconf,authorversion]{acmart}

\AtBeginDocument{%
  \providecommand\BibTeX{{%
    \normalfont B\kern-0.5em{\scshape i\kern-0.25em b}\kern-0.8em\TeX}}}

\setcopyright{acmcopyright}
\copyrightyear{2020}
\acmYear{2020}
\acmDOI{10.1145/1122445.1122456}

\acmConference[KDD Workshop on Machine Learning in Finance '20]{KDD Workshop on Machine Learning in Finance '20: KDD Workshop on Machine Learning in Finance}{August 24, 2020}{}
\acmBooktitle{KDD Workshop on Machine Learning in Finance '20,
  August 24, 2020}
\acmPrice{15.00}
\acmISBN{978-1-4503-XXXX-X/18/06}

\begin{document}

\title{FairXGBoost: Fairness-aware Classification in XGBoost}

%
\author{Srinivasan Ravichandran}
\email{srinivasan.ravichandran@aexp.com}
\affiliation{%
  \institution{AI Labs, American Express}
  \city{Bangalore}
  \state{Karnataka}
}
\author{Drona Khurana}
\email{drona.khurana@aexp.com}
\affiliation{%
  \institution{AI Labs, American Express}
  \city{Bangalore}
  \state{Karnataka}
  }
  
  \author{Bharath Venkatesh}
\email{bharath.venkatesh@aexp.com}
\affiliation{%
  \institution{AI Labs, American Express}
  \city{Bangalore}
  \state{Karnataka}
  }

\author{Narayanan Unny Edakunni}
\email{narayanan.u.edakunni@aexp.com}
\affiliation{%
  \institution{AI Labs, American Express}
  \city{Bangalore}
  \state{Karnataka}
}









\begin{abstract}
 Highly regulated domains such as finance have long favoured the use of machine learning algorithms that are scalable, transparent, robust and yield better performance. One of the most prominent examples of such an algorithm is XGBoost\cite{chen2016xgboost}. Meanwhile, there is also a growing interest in building fair and unbiased models in these regulated domains and  numerous bias-mitigation algorithms have been proposed to this end. However, most of these bias-mitigation methods are restricted to specific model families such as logistic regression or support vector machine models, thus leaving modelers with a difficult decision of choosing between fairness from the bias-mitigation algorithms and scalability, transparency, performance from algorithms such as XGBoost. We aim to leverage the best of both worlds by proposing a fair variant of XGBoost that enjoys all the advantages of XGBoost, while also matching the levels of fairness from the state-of-the-art bias-mitigation algorithms. Furthermore, the proposed solution requires very little in terms of changes to the original XGBoost library, thus making it easy for adoption. We provide an empirical analysis of our proposed method on standard benchmark datasets used in the fairness community.
\end{abstract}




\maketitle

\section{Introduction}
Machine learning models are increasingly replacing traditional modeling systems because of better predictive performance and scalability. This has resulted in an explosion of the number of machine learning models being used in decision-making systems for a broad spectrum of activities such as credit lending, candidate recruitment etc. This high rate of adoption also means that machine learning models have a significant impact on people and the society at large. Consequently, it is essential to ensure that these models are well-regulated.

Building machine learning models in highly regulated domains such as finance, healthcare etc. often comes with its own share of additional challenges. For instance, in the finance industry, it is essential to be transparent in the decision-making process. These challenges can also arise in the form of legal requirements such as the Equal Credit Opportunity Act (ECOA)\cite{ECOA} which makes it unlawful for any creditor to discriminate against any applicant on the basis of race, sex, color etc. 

Regulatory bodies all around the world such as European Union through its General Data Protection and Regulation (GDPR) \cite{GDPR} have been keen on developing ways to make machine learning models fair, accountable, transparent and explainable. A host of bias-mitigation approaches \cite{calmon2017optimized}\cite{FATB}\cite{hardt2016equality}\cite{KamishimaAAS12} have been proposed in recent times that ensure that models are not discriminatory against a specific population. While most of the initial work involved ensuring fairness in the training data being used to train these models, these methods often suffered a loss in model performance. More sophisticated bias mitigation strategies\cite{Zafar}\cite{zhang2018mitigating} have been proposed recently and have been demonstrated to be effective in ensuring fairness, while also not compromising too much on performance. However, most of these bias mitigation strategies are designed for specific classes of models such as neural networks, which often violate the regulatory requirement of being transparent or for models such as logistic regression and support vector machines, which are typically inferior in performance and scalability. Thus, on the one hand, we have simple, interpretable and high-performing machine learning algorithms such as XGBoost that are widely preferred in the finance industry and on the other hand, we have sophisticated bias mitigation strategies that are designed to work with a completely different family of machine learning algorithms such as neural networks which may not be suitable for usage in regulated domains.

Our goal is to bridge this gap by formulating equivalent bias mitigation strategies for more practical algorithms. Specifically, in this paper, we introduce a bias-mitigation scheme for XGBoost \cite{chen2016xgboost}. XGBoost has the desirable advantages of being flexible, explainable and scalable while also providing state-of-the-art performance in most supervised learning tasks. However, XGBoost differs significantly from other convex-margin based classifiers such as Support Vector Machines (SVM), logistic regression, neural networks etc. in the way in which the model parameters are updated. This is the primary reason why existing bias-mitigation methods are not compatible with XGBoost.

Our main contributions in this paper are 
\begin{itemize}
    \item the formulation of a regularization-based in-process bias mitigation technique that squarely fits into XGBoost's greedy tree building algorithm
    \item empirical comparison with other state-of-the-art bias mitigation strategies on common benchmarks typically used in the bias-mitigation literature 
\end{itemize} 

The paper is structured as follows. In Section 2, we provide a brief overview of the existing state-of-the-art bias-mitigation methods. In Section 3, we define our bias-mitigation framework, derive gradient and hessian values for model building and show how it fits into XGBoost's existing framework and requires very little in terms of modification. In Section 4, we show our experimental results on common benchmark datasets in the fairness literature and compare our framework's performance and fairness against state-of-the-art bias mitigation strategies.

\section{Preliminaries}
As noted in the introduction, bias-mitigation is a growing area of research, thanks to the increasing interest in machine learning model regulation. Typically, these regulations prohibit discrimination in decision-making against a certain population characterized by \textit{sensitive attributes} such as race, gender, age, marital status etc. While one might be tempted to assume that excluding these sensitive attributes in the model building process (which is a common practice in the industry) would result in fair models, \cite{disc-aware} demonstrated that this is not always the case. Often times, other proxy variables that are correlated with these sensitive attributes carry sufficient information to induce bias in the model. Additionally, the datasets used to train these models and the processes that generate these datasets might themselves be inherently biased. Hence, we need a sophisticated bias-mitigation algorithm to overcome this problem.

\subsection{Notation and Metrics}
In this paper, we consider a supervised learning task and we adopt the following notation. The model is trained using $n$ training samples denoted by $(x_i, s_i, y_i)_{i=1}^{n}$. For the $i^{th}$ training sample, $x_i$ is the feature vector , ${s_i}$ is the binary indicator for the sensitive attribute (such as sex or race), $y_i$ is the binary target label and $\hat{y_i}$ is the score produced by the model. Note that in the case of binary classification, $\hat{y_i}$ needs to be transformed using the sigmoid function before being interpreted as the output probabilities for the two classes. In order to quantify the level of bias exhibited by the model, we need a fairness metric. Multiple fairness metrics have been proposed and there isn't a one-size-fits-all metric. In this paper, we will be focusing on the following metric.

\begin{definition}
The disparate impact for a model is the ratio of the positive prediction rates of the minority and the majority groups.
\begin{align}
    DI = \frac{P(\hat{Y} = 1 | S = 0)}{P(\hat{Y} = 1 | S = 1)}
\end{align}
\end{definition}
Disparate impact (DI) is a well-established measure of fairness and is often associated with the 80\% rule that is frequently cited in legislation. We consider this metric due to its prevalence in the regulatory law. Throughout this paper, we will be using DI as the metric for fairness and accuracy as the metric for measuring model performance.
\subsection{Prior Work}Prior work on bias mitigation can be broadly classified into three categories: pre-processing, in-processing and post-processing methods. Pre-processing methods \cite{calmon2017optimized}\cite{kamiran2012data} typically project the data into a feature space with fair representations. In-processing methods \cite{FATB}\cite{KamishimaAAS12}\cite{zhang2018mitigating} involve changing the training procedure in order to make the model predictions fair. Post-processing \cite{hardt2016equality}\cite{kamiran2012decision} methods typically transform the model outputs to ensure fairness. Of these three categories, in-processing methods offer maximum robustness and flexibility. Existing in-processing methods can further be classified into four categories: optimization in a space constrained by a fairness metric \cite{Zafar}, a regularized objective function on an unconstrained space where the regularizer is typically a function of the model output and the sensitive feature \cite{KamishimaAAS12}, an adversarial learning set-up where an adversary attempts to identify the correlation between a sensitive attribute and the predictor model's output while the predictor model's goal is to maximize performance and simultaneously fooling the adversary \cite{zhang2018mitigating} and finally, designing meta-algorithms \cite{celis2019classification}.

Both pre-processing and post-processing methods have been widely applied for black-box models. However they are often inflexible and result in degradation of model performance. In-processing methods on the other hand provide robust bias mitigation with a relatively lower performance degradation. \cite{Zafar} proposed an in-process method where the search space for the model is constrained by a fairness metric, namely the co-variance between the sensitive features and the signed distance of the instance from the model's decision boundary. However, their method is applicable only to the family of convex-margin classifiers and not to algorithms such as XGBoost. Kamishima et al.\cite{KamishimaAAS12} proposed the prejudice remover which is a regularization based bias-mitigation strategy. The idea is to add a regularizer that captures the mutual information between $Y$ and $S$. Once again, their method cannot be extended to algorithms such as XGBoost. \cite{zhang2018mitigating} proposed an adversarial setup as described earlier. However, their method suffers from poor convergence characteristics and it is often difficult to tune the adversarial system.

Literature on fairness in ensemble models, especially in boosted tree models is rather limited. To the best of our knowledge, only \cite{fish2015fair} and \cite{FATB} consider fairness in a boosting setup. \cite{fish2015fair} were the first to perform a case study of fairness for Adaboost. Their approach involved pre-processing and post-processing methods such as random reshuffling which can incur additional performance degradation. \cite{FATB} proposed fair adversarial gradient tree boosting where the predictor from \cite{zhang2018mitigating} was a decision tree model. However, the issues of convergence from an adversarial setup still remain, rendering the method to be often impractical.

\subsection{Gradient Boosted Decision Trees}
Gradient Boosted Decision Trees (GBDT), introduced by \cite{friedman2001greedy}, is a boosting framework consisting of a collection of weak learners which are shallow decision trees. \cite{chen2016xgboost} proposed eXtreme Gradient Boosting (XGBoost) as a scalable end-to-end tree boosting algorithm. XGBoost has enjoyed widespread adoption by data scientists in the machine learning community. XGBoost has gained particular interest in finance owing to the fact that it is flexible, scalable and explainable.

The GBDT setup for a dataset $D = \{(x_i, y_i)\}$ involves $K$ additive functions put together to make a prediction. Formally, a GBDT model consists of $K$ trees each represented as $f_t(x)$ built at the $t^{th}$ boosting round. The prediction function is then defined as 
\begin{align}
    \hat{y_i} =\sum_{t=1}^{K} f_t(x_i) \nonumber
\end{align}

The trees are built in a greedy manner by optimizing the following objective function
\begin{align}
    \mathcal{L}^{t} = \sum_{i=1}^{n} l(y_i, \hat{y}^{(t-1)}_i + f_t(x)) + \Omega(f_t) 
\end{align}
where $l$ is an appropriate loss function that depends on the task at hand and $\Omega$ is a regularizer for the tree structure. For classfication tasks, a common choice is the cross-entropy loss between $y_i$ and $\hat{y_i}$, while the squared error is used for regression tasks. 

The key contribution of \cite{chen2016xgboost} is the reformulation of this optimization problem as follows. The objective function in the above equation can be approximated using the Taylor expansion as 
\begin{align}
    \mathcal{L}^{t} = \sum_{i=1}^{n} l(y_i, \hat{y}^{(t-1)}_i) + g_if_t(x_i) + \frac{1}{2}h_if_t(x_i)^2 + \Omega(f_t)
\end{align}
where $g_i = \nabla_{\hat{y}^{(t-1)}_i} l(y_i, \hat{y}^{(t-1)}_i)$ and $h_i = \nabla^2_{\hat{y}^{(t-1)}_i} l(y_i, \hat{y}^{(t-1)}_i)$. This objective function is then transformed from the space of $f_t$ to the space of node weights $w_j$ of the trees, which results in the following.
\begin{equation}
    \mathcal{L}^{t} = \sum_{j=1}^{T}\left[w_j\left(\Sigma_{i\in I_j}g_i\right) + \frac{1}{2}w_j^2\left(\Sigma_{i\in I_j} h_j\right)\right] + \Omega(f_t)
\end{equation}
where $I_j$ is the set of indices of the samples that fall in the leaf $j$. The best split is computed as the split value that optimizes this objective function.

\section{Proposed Framework}
Our proposed approach involves the use of a fairness regularizer that aims to remove correlation between the sensitive attribute and the target value, thereby ensuring model fairness. The extent to which the regularizer affects the model is controlled by a hyperparameter.

\subsection{Fairness Regularizer}
Using the notation that we introduced earlier, we have a set of training samples $D = \{(x_i, s_i, y_i)\}$. For convenience, let us assume, without loss of generality $s_i = 1$ represents instance $i$ belonging to the majority group and $s_i = 0 $ represents instance $i$ belonging to the minority group. Similarly, $y_i = 1$ represents a \textit{favourable outcome} (such as approval of a credit application). Let $\hat{y_i}$ be the raw leaf score produced by the model for the $i^{th}$ instance and  $\sigma(z) = \frac{1}{1+e^{-z}}$ be the classic sigmoid function. We propose the following regularizer. 
\begin{align}
\mathcal{R}^t = \sum_{i=1}^{n} -s_i\  log\left(\sigma(\hat{y}^{(t)}_i)\right) - (1-s_i) \ log\left(1 - \sigma(\hat{y}^{(t)}_i)\right)
\end{align} 

We emphasise that one must choose the encoding for the majority and minority population as follows. If $t \in \{0,1\}$ represents the favourable outcome in a classification task (for example, $t = 0$ in a credit risk model if the outcome $0$ corresponds to a customer being classified as low-risk), then the members of the minority group should be encoded with $s = t$ and the majority members must be encoded with $s = 1-t$. Intuitively, this encoding enables the regularizer to push for more favourable outcomes to the minority group and leads to a decrease in the bias of the model.
\subsection{Gradient and Hessian for the Regularized Objective}
The regularized objective function for a supervised classification task will now be the sum of the  classical cross-entropy loss between the model predictions and the ground truth labels and the negative cross-entropy between the model predictions and the sensitive feature. 
\begin{align}
    \bar{\mathcal{L}^{t}} = &\sum_{i=1}^{n} - y_i\  log\left(\sigma(\hat{y}^{(t)}_i)\right) - (1-y_i) \ log\left(1 - \sigma(\hat{y}^{(t)}_i)\right) + \Omega(f_t) \  \nonumber \\
    &- \mu \sum_{i=1}^{n} s_i\  log\left(\sigma(\hat{y}^{(t)}_i)\right) + (1-s_i) \ log\left(1 - \sigma(\hat{y}^{(t)}_i)\right) \nonumber
\end{align}
The hyperparameter $\mu$ determines the strength of the regularizer: the higher the regularizer strength the higher the fairness score of the model. This gives us fine-grained control over the level of fairness we desire. It should be noted that the choice of $\mu$ should be such that $\mu \geq 0$, in order to avoid unboundedness in the direction of optimization.

We re-trace the steps of \cite{chen2016xgboost} and reformulate this objective from the space of functions $f_t$ to the space of node weights $w_j$, by computing the gradient $\bar{g}_i$ and hessian $\bar{h}_i$ for the new objective function as follows. We drop the superscript $(t-1)$ for convenience of notation. 

\begin{align}
    \bar{g}_i =\  &\nabla_{\hat{y}_i} \left(\sum_{i=1}^{n} - y_i\  log\left(\sigma\left(\hat{y}_i\right)\right) - (1-y_i) \ log\left(1 - \sigma\left(\hat{y}_i\right)\right)\right) \nonumber \\ 
     &\ + \ \mu\ \nabla_{\hat{y}_i} \left(\sum_{i=1}^{n} s_i\  log\left(\sigma\left(\hat{y}_i\right)\right) + (1-s_i) \ log\left(1 - \sigma\left(\hat{y}_i\right)\right)\right) \nonumber \\ 
    \bar{g}_i =\  & \sigma\left(\hat{y}_i\right) - y_i + \mu \left(s_i - \sigma\left(\hat{y}_i\right) \right) \nonumber
\end{align}
Similarly, we can derive $\bar{h}_i$ and we obtain
\begin{align}
\bar{h}_i =\  &\nabla^2_{\hat{y}_i} \left(\sum_{i=1}^{n} - y_i\  log\left(\sigma\left(\hat{y}_i\right)\right) - (1-y_i) \ log\left(1 - \sigma\left(\hat{y}_i\right)\right)\right) \nonumber \\ 
     &\ + \ \mu\ \nabla^2_{\hat{y}_i} \left(\sum_{i=1}^{n} s_i\  log\left(\sigma\left(\hat{y}_i\right)\right) + (1-s_i) \ log\left(1 - \sigma\left(\hat{y}_i\right)\right)\right) \nonumber \\ 
    \bar{h}_i=\  & (1-\mu)\sigma\left(\hat{y}_i\right) \left(1 - \sigma\left(\hat{y}_i\right)\right) \nonumber 
\end{align}

The rest of the tree building process remains the same as XGBoost except that we use the new $\bar{g}_i$ and $\bar{h}_i$ instead. It is worth noting that comparing $\bar{g}_i$ and $\bar{h}_i$ with $g_i$ and $h_i$ from the original XGBoost formulation, we get the following relationships for the gradient and hessian.
\begin{align}
    \bar{g}_i\ =\  &g_i + \mu \left(s_i - \sigma\left(\hat{y}_i\right) \right) \\
    \bar{h}_i\ =\  &h_i \left(1 - \mu \right)
\end{align}

This simple relationship between the original $g_i, h_i$ and our proposed $\bar{g}_i, \bar{h}_i$ is what makes our approach appealing since it can be directly implemented using the custom objective feature of XGBoost. In the next section, we describe the experimental setup that we used and compare our approach to the current state-of-the art methods. Additionally, we also provide insights on how the fairness of the model changes as we increase $\mu$.

\section{Experimental setup}

Throughout this section, we take the following approach. The best hyperparameter settings for the XGBoost model such as \texttt{max-depth}, \texttt{num-rounds}, \texttt{learning-rate} have been identified as the ones that maximize model accuracy, without the fairness regularizer in place ($\mu = 0$). The same setting of hyperparameters is then used and models are built with different values of $\mu$ and the corresponding accuracy and fairness metrics are measured and reported.

The fairness metric we report is the disparate impact (DI) defined in Section 2. The datasets we use are the standard benchmark datasets that are used in the bias-mitigation literature, which are described below. 

The first dataset is the UCI Adult Income dataset \cite{Dua:2019} where the goal is to train a model that can predict if an individual makes more than \$50K as income, given a set of features such as age, capital gains and capital losses. The dataset also contains the sensitive attribute \texttt{sex} which takes on two values \texttt{\{Male, Female\}}. The dataset comprises of more males than females, thus making \texttt{Male} the majority population. Fairness here would imply that the model predictions do not discriminate against \texttt{Female}.

The second dataset that we consider is the COMPAS recidivism dataset \cite{COMPAS}. The model is trained to predict if an individual is likely to re-offend in the future. The dataset contains 13 features about 7000 individuals. This was one of the hallmark datasets that was used in the first major debate on the fairness of machine learning models. The sensitive attribute being considered here is \texttt{race}. Once again, fairness here would mean that no particular race is discriminated against.

In addition to the above, we also consider the two other datasets that were analyzed in \cite{FATB} namely the Bank and the Default datasets. The Default dataset \cite{taiwancc} comprises of 23 features about 30,000 Taiwanese credit card users with class
labels which state whether an individual will default on
payments. The sensitive attribute that is being considered for this dataset is \texttt{sex}.

The Bank dataset \cite{bankdata} consists of 16 features of about 45000 clients of a Portuguese banking institution. The goal of the task is to predict if the client has subscribed to a term deposit. The sensitive attribute is \texttt{age} after it has been encoded in a binary format indicating if a customer is between 33 and 60 years old or not.

\begin{table}
  \caption{Benchmark dataset statistics}
  \label{tab:freq}
  \begin{tabular}{crc}
    \toprule
    Name & Number of rows&Sensitive Attribute\\
    \midrule
    Adult & $\sim32000$ & sex\\
    COMPAS & $\sim7000$ & race\\
    Default & $\sim30000$ & sex\\
    Bank & $\sim45000$ & age\\
  \bottomrule
\end{tabular}
\end{table}

\begin{table}
  \caption{Drop in accuracy to achieve DI $\geq 80\%$}
  \label{tab:di_stats}
  \begin{tabular}{ccccc}
    \toprule
    Dataset & FairXGB & Grari\cite{FATB} & Kamishima\cite{KamishimaAAS12} & Zhang\cite{zhang2018mitigating}\\
    \midrule
    Adult & 4.4\% & \textbf{1.9}\% & 3.0\% & 2.3\%\\
    COMPAS & \textbf{1.0}\% & 4.6\% & 3.7\% & 1.0\%\\
    Default & \textbf{0.0}\% & 0.7\% & 1.0\% & \textbf{0.0}\%\\
    Bank & \textbf{0.5}\% & 0.6\% & 0.7\% & 0.6\%\\
  \bottomrule
\end{tabular}
\end{table}




We compare our work to three in-process bias mitigation methods - prejudice remover \cite{KamishimaAAS12}, the fair adversarial gradient tree boosting \cite{FATB} and adversarial debiasing \cite{zhang2018mitigating}. For prejudice remover, we use the implementation provided by \cite{KamishimaAAS12} for training the model. For \cite{FATB}, we re-use the numbers reported by them as the benchmark since the hyperparameters are unknown and hence we cannot reproduce their results. Since different model families would provide different accuracies to begin with, we measure the drop in accuracy from the vanilla model in order to obtain a DI of at least 80\%, rather than the absolute value of the accuracy itself. The 80\% DI target is arbitrary and is often a useful rule of thumb that comes from the 80-20 rule\cite{EEOC}. The comparisons in the drop in accuracy incurred by our approach against \cite{FATB}, \cite{KamishimaAAS12} and \cite{zhang2018mitigating} is shown in Table \ref{tab:di_stats}.

Our method outperforms \cite{FATB}, \cite{zhang2018mitigating} and \cite{KamishimaAAS12} on all but the Adult dataset, where the drop in accuracy is more pronounced. When compared against \cite{KamishimaAAS12}, our method and \cite{FATB} incur a smaller dip in accuracy, thus showing their effectiveness. An interesting observation is that for the Default dataset, the vanilla XGBoost model that we trained was already satisfying the critera of $DI \geq 80\%$ before any bias mitigation was applied. The same was observed in the case of the vanilla neural network model for \cite{zhang2018mitigating}. We believe that the higher loss in accuracy for the Adult dataset could be explained by low DI of the vanilla model. Thus, the best hyperparameters for the vanilla model need not necessarily be optimal for all values of $\mu$. This is supported by the fact that the adversarial methods \cite{FATB} and \cite{zhang2018mitigating} incur a lower loss in accuracy for the Adult dataset because they use a more complex multi-layer perceptron adversary. A more nuanced method for tuning hyperparameters is hence required for non-adversarial methods and we defer this to future work.

\begin{figure}
  \includegraphics[width=\columnwidth]{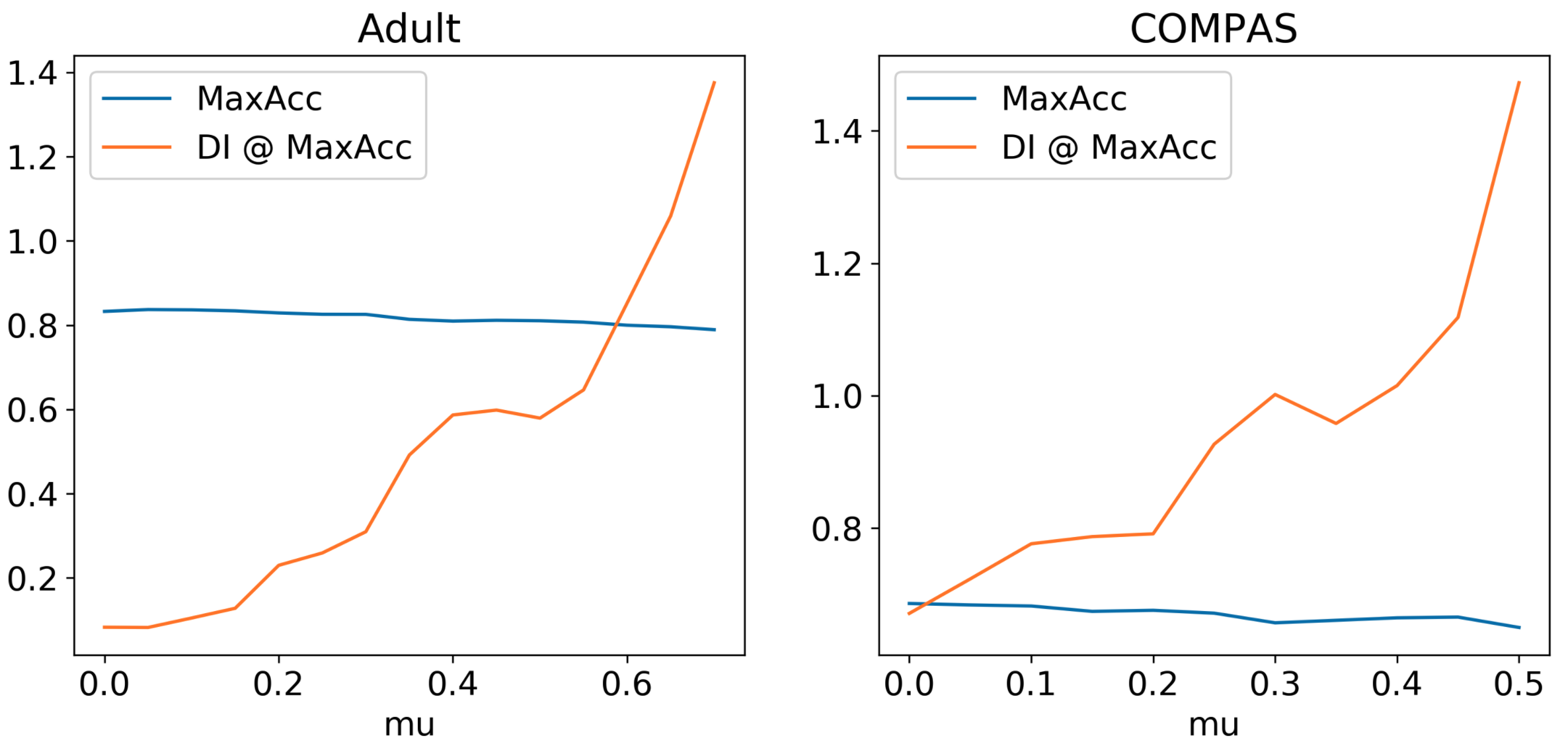}
  \caption{Disparate Impact of the maximum-accuracy classifier for different values of $\mu$ for the Adult Income Dataset and the COMPAS dataset}
  \Description{}
    \label{fig:adult_compas}

\end{figure}

\begin{figure}
  
  \includegraphics[width=\linewidth]{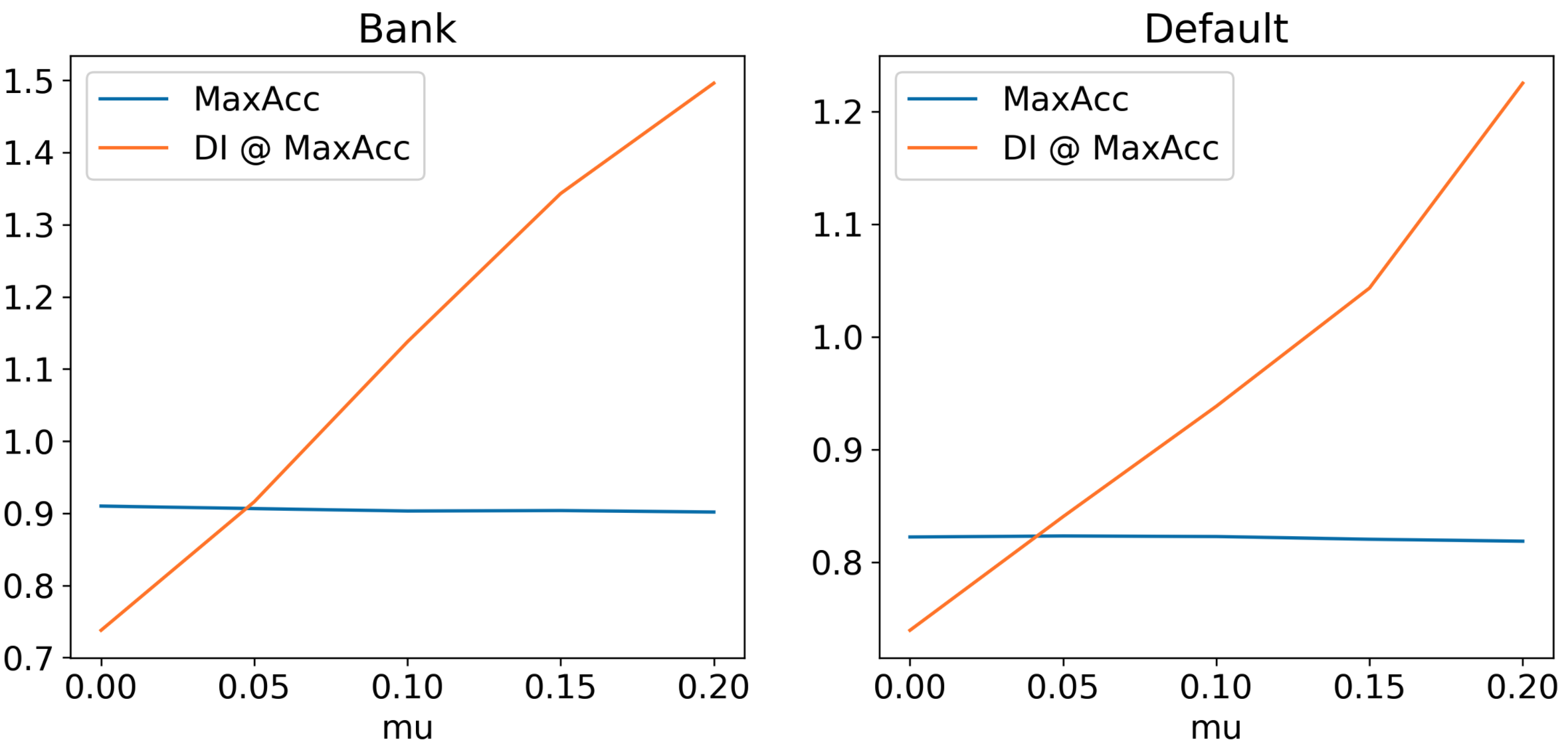}
  \caption{Disparate Impact of the maximum-accuracy classifier for different values of $\mu$ for the Bank dataset and the Default dataset}
  \Description{DI vs thresh}
  \label{fig:bank_default}
\end{figure}

We also plot the variation of DI and accuracy with respect to the weight of the fairness regularizer $\mu$, to visualise the effect of increasing $\mu$ on the DI and accuracy metrics. For each $\mu$, we pick the classifier that achieves the highest accuracy and report it along with the corresponding disparate impact. This is in contrast to some of the prior analyses where the "best" models are chosen as those with the best DI. We believe that studying the fairness of the best-accuracy model is more practical for modelers. In Fig. \ref{fig:adult_compas}, we show the plot for the Adult and COMPAS datasets. The Adult dataset requires a weight in the range between $\mu=0.6$ and $\mu=0.7$ to reach the acceptable range of DI which is anything greater than 0.8, whereas the COMPAS dataset reaches the acceptable range at a lower value of $\mu$, between $\mu=0.2$ and $\mu=0.5$. Similarly, the plots for the Bank and Default datasets are shown in Fig.\ref{fig:bank_default}.

\section{Conclusion}
In this paper, we have described an extension of XGBoost that can be used to build fair machine learning models. Our choice of the regularizer for fairness makes it easy to be incorporated into XGBoost with minimal changes, while also providing fine-grained control of the level of fairness that needs to be imposed. Furthermore, we have compared our method with the current state-of-the-art bias mitigation strategies on common benchmark datasets. While we have only considered the cross-entropy loss between $\hat{y}_i$ and $s_i$, our framework is applicable to other continuous and differentiable regularizer functions as well. Hence, our proposal helps bridge the gap between fairness researchers and practitioners in the finance community.

As future work it would be interesting to tackle the other challenge that is typically faced in well-regulated domains - privacy. There have been methods such as differential privacy that have been proposed for secure sharing of sensitive features to modelers to build their models. Adopting such a methodology for XGBoost would be a good addition that would go a long way. Another interesting direction to pursue would be the monitoring of the regularized objective in order to gain insights on the fairness-accuracy tradeoffs. XGBoost's inherent support to track an evaluation metric could be reused for this task.

It would also be useful to pursue the handling of polyvalent sensitive attributes (such as race which can take on many values such as Asian, White, Hispanic, African-American etc.).




\bibliographystyle{ACM-Reference-Format}
\bibliography{kdd-mlf-fairxgb.bbl}










\end{document}